\documentclass{article}
\usepackage{titling}
\usepackage{arxiv}
\usepackage{times}
\usepackage{latexsym}
\usepackage{amsmath}
\usepackage{amssymb}
\usepackage{amsfonts}
\usepackage{booktabs}
\usepackage{threeparttable}
\usepackage{url}
\usepackage{graphicx}
\usepackage{epsfig}
\usepackage{tikz}
\usepackage{cite}
\usepackage{subfigure}
\usepackage{multirow}
\usepackage{caption}

\graphicspath{{./figures/}}

\newcommand\mybox[2][]{\tikz[overlay]\node[fill=blue!20, inner sep=0pt, minimum height=4mm, anchor=text, rectangle, rounded corners=1mm,fill=red!#1] {#2};\phantom{#2}}

\title{Combine Convolution with Recurrent Networks for Text Classification}

\author{
	Shengfei Lyu \\
	School of Computer Science and Technique\\
	University of Science and Technique of China\\
	Hefei, Anhui 230027 \\
	\texttt{saintfe@mail.ustc.edu.cn} \\
	\And
	Jiaqi Liu \\
	School of Computer Science and Technique\\
	University of Science and Technique of China\\
	Hefei, Anhui 230027 \\
	\texttt{jiaqiliu@mail.ustc.edu.cn} \\
}

\date{}
\begin{document}
\maketitle
\begin{abstract}

Convolutional neural network (CNN) and recurrent neural network
(RNN) are two popular architectures used in text classification.
Traditional methods to combine the strengths of the two networks
rely on streamlining them or concatenating features extracted from
them. In this paper, we propose a novel method to keep the strengths
of the two networks to a great extent. In the proposed model, a
convolutional neural network is applied to learn a 2D weight matrix
where each row reflects the importance of each word from different
aspects. Meanwhile, we use a bi-directional RNN to process each word
and employ a neural tensor layer that fuses forward and backward
hidden states to get word representations. In the end, the weight
matrix and word representations are combined to obtain the
representation in a 2D matrix form for the text. We carry out
experiments on a number of datasets for text classification. The
experimental results confirm the effectiveness of the proposed
method.

\end{abstract}


\section{Introduction}

Text classification is one of the fundamental tasks in Natural
Language Processing (NLP). The goal is to classify texts according
to certain criteria. It has many practical applications such as
sentiment analysis \cite{pang2005seeing} and topic categorization
\cite{wang2012baselines}.


Traditional approaches, for example, bag-of-words, extract features
such as statistics on unigrams and bigrams for a general classifier.
In recent years, the development of pre-trained word embeddings and
deep neural networks has brought new inspiration to various NLP
tasks. Word embeddings are used to map words to an implicit space
where semantic relationships are preserved in their reciprocal
closeness commonly measured by the Euclidean norm. This type of
representation can alleviate the data sparsity problem
\cite{bengio2003neural}. Moreover, researchers demonstrate that
pre-trained word embeddings are able to capture meaningful syntactic
and semantic regularities
\cite{pennington2014glove,xu-etal-2016-improve,Liu2017revisit}. With
the help of word embeddings, deep learning models such as
convolutional neural network (CNN) \cite{lecun1998gradient} and
recurrent neural network (RNN) \cite{elman1990finding} are proposed
to further process the semantic representation within texts. This
methodology has made impressive progress in text classification
\cite{kalchbrenner2014convolutional,Jiang2018LatentTT}.

CNN has been proven to be a powerful semantic composition model for modeling
texts \cite{kim2014convolutional}. CNN treats
texts as a 2D matrix by concatenating embedding of words together. It utilizes
a 1D convolution operator to perform the feature mapping, and then conducts a 1D
pooling operation over the time domain for obtaining a fixed-length output
feature vector. Based on the convolution operation, it is able to capture both
local and position-invariant features in texts.  Alternative popular neural
network model, RNN \cite{hochreiter1997long}, treats texts as sequential data
and analyzes texts word by word.  Fixed-length hidden units in RNN stores
the contextual information up to the present word. Long Short-term Memory (LSTM)
units \cite{hochreiter1997long} and gated recurrent units (GRU)
\cite{cho2014learning} are two popular prototypes that aim to solve gradient
vanishing and gradient explosion problems.

Some approaches attempt to combine CNN and RNN to incorporate the
advantages of both models
\cite{lai2015recurrent,wang2017hybrid}.
However, most of them integrate CNN and RNN only by streamlining the
two networks \cite{lai2015recurrent,
 wang2017hybrid}, which might
decrease the performance of them.
 \begin{figure}[!tbp]
    \centering
    \caption{Examples of topic classification   }
    \fbox{
        \begin{minipage}{0.95\linewidth}
            \begin{itemize}
                \item A new home heating system from \textbf{Panasonic} is based on a \textbf{hydrogen fuel cell}; it both heats the house and produces hot water.
                \item \textbf{Skype} is the easiest, fastest and cheapest way for individual customers to use their computers with \textbf{broadband connections} as telephones.

            \end{itemize}
        \end{minipage}
    }
    \label{topic_classify}
 \end{figure}
 In addition, prior works neglect the fusion of contextual information when learning the word's contextual representation.
 Most methods incorporating bi-directional RNN to model each word's contextual representation usually choose concatenating the forward and backward hidden states at each time step \cite{tang2015document,  wang2017hybrid}.
 As a result, the resulting vector does not have interaction information of the forward and backward hidden states.
 Meanwhile, the hidden state in one direction may be ``myopic'' and against the meaning collected by another hidden state.
 Intuitively, a word’s contextual representation is more accurate when holistic semantics are
 collected and fused from two directions.
 Failure in doing so may lose true meaning for the focus word,
 especially for polysemic words whose meanings are context-sensitive.

In this paper, we propose a neural network model that incorporates
CNN and RNN in a novel way. The intuition underlying our model is
that different words contribute differently to the meaning of the
whole text and the key parts can be well extracted by CNN. As an
example in Figure \ref{topic_classify}, the words in bold are the
most informative for the sentences labeled as \textit{Science and
Technology}. Due to the capacity of capturing local and
position-invariant features, CNN is utilized to extract local
features and learn a 2D matrix that shows the importance of each
word from different aspects. Meanwhile, the bi-directional RNN is
applied to learn  word contextual representations. A neural tensor
layer is introduced on the top of the bi-directional RNN to obtain
the fusion of the bi-directional contextual information surrounding
the focus word. We call this novel neural network as convolutional
recurrent neural network (CRNN) and apply it to the task of text
classification. By combining the convolution and recurrent
structure, our method preserves the advantage of both CNN and RNN.

Our contributions in this paper are listed as follows:

\begin{itemize}
    \item A convolutional neural network is proposed to obtain relative importance of each word in the text, which not only could improve model performance but also help to interpret classification results.

    \item A neural tensor layer is added on the top of bi-directional RNN to obtain the representation of each word. Compared with direct concatenating forward and backward hidden states, this way allows the  fusion of bi-directional contextual information surrounding the focus word.

    \item  We present a novel architecture neural network called CRNN and experiments on six
    real-world datasets in different text classification tasks. The results demonstrate that the proposed approach outperforms
    all the selected baselines. Furthermore, it could offer the importance information of each word for the prediction result.
\end{itemize}

\section{Method}

\begin{figure*}[!htbp]
    \centering \includegraphics[width = \textwidth]{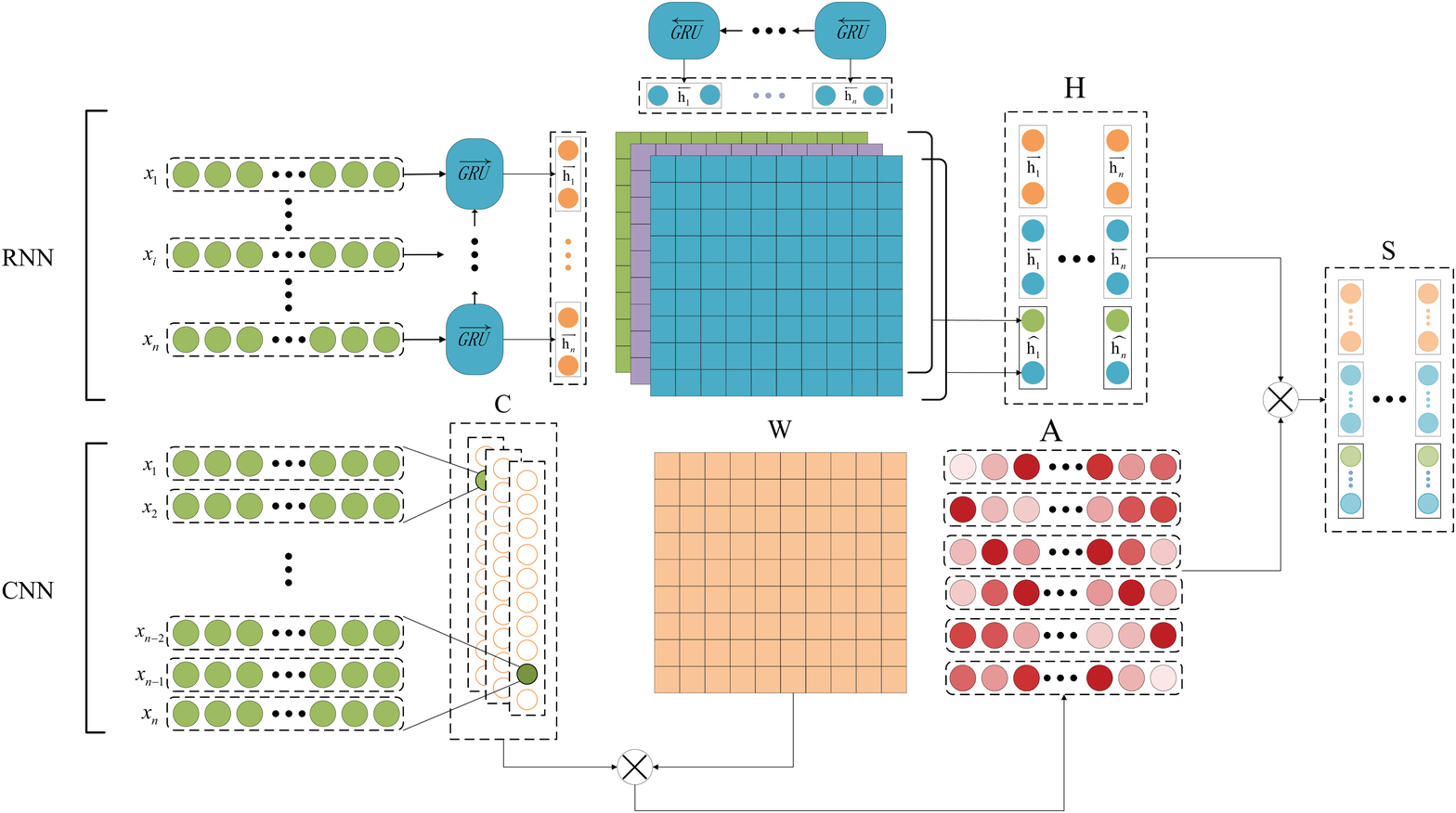}
    \setcaptionwidth{\textwidth} \caption{The architecture of CRNN. The text
        ${X}$ serves as input for CNN and RNN. We apply a convolution operation on
        matrix ${\mathbf{X}}$ to learn a 2D weight matrix, denoted by $\textbf{A}$.
        Each row in the matrix captures semantics of the text and their information
        should not be redundant. Meanwhile, a bi-directional RNN is employed to extract
        the contextual features of words. Based on bi-directional feature vectors, a
        neural tensor layer fuses information from two directions and obtains matrix
        $\textbf{H}$. Finally, $\textbf{A}$ and $\textbf{H}$ are multiplied to get
        the representation of the text, denoted by matrix $\textbf{S}$. } 
    \label{CRNN}
\end{figure*}

In this section, we introduce the proposed method, CRNN, which uses
a CNN to compute the importance of each word in the text and
utilizes a neural tensor layer to fuse forward and backward hidden
states of  bi-directional RNN. Fig. \ref{CRNN} shows the
architecture of CRNN. The input of the network is a text denoted by
$X$, which is a sequence of words ${x_1},{x_2}, \ldots ,{x_n}$. The
output of the network is the text representation that is
subsequently used as input of a fully-connected layer to obtain the
class prediction. We will introduce the details of the main
components in the following sections. First, we explain how to use
CNN to compute the importance of each word in a text. Second, a
bi-directional RNN followed by a neural tensor layer is described.
Finally, we introduce the training details and regularization in our
method.

 \subsection{Weight matrix learning}
Usually, the text contains more than one focus word, like a phase or a group of related words.
The focus words together form  the overall semantic meaning of the sentence.
Learning weight matrix of words not only could lead to better performance, but also provides insight into which words
and sentences contribute to the classification decision, which can be of value in applications and analysis\cite{shen2014latent,gao2014modeling}.
In this part, we illustrate how to use CNN to extract the importance of each word.

 Given a text as a sequence with $n$ words $X = [{x_1},{x_2}, \ldots ,{x_n}]$. Each word ${x_i}$ in the text is denoted by a $d$-dimensional word vector ${{\mathbf{x}}_i}$. The text is represented as a matrix ${\mathbf{X}} \in {\mathbb{R}^{n\times d}}$ by concatenating each word vector. The convolution operation is performed with a $k$-size window filter ${{\mathbf{W}}_j} \in {\mathbb{R}^{k\times d}}$ on every $k$ consecutive word embeddings:
 \begin{equation}
 {{c_{ji}}} = f({{\mathbf{W}}_j}\cdot{{\mathbf{X}}_{i:i + k - 1}} + b),
 \end{equation}
 where $b \in R$ is a bias term and the operator $\cdot$ means dot product. We adopt ReLU  \cite{nair2010rectified} for $f$
 in this study and use a colon to represent a continuous range of indexes. The result $c_{ji}$ represents the feature
  of the $j$-th $k$-gram  extracted by filter $\mathbf{W}_j$.
 The $ j $-th filter ${{\mathbf{W}}_j}$ traverses across the word embedding sequence along the temporal axis.
 We pad $\left\lceil {(k - 1)/2} \right\rceil$ and $\left\lfloor
 {(k - 1)/2} \right\rfloor $ $d$-dimensional zero vectors at the beginning and end
 of ${{\mathbf{X}}_{1:n}}$.
 The local view is concatenated in the output feature vector ${\mathbf{c_{j}}}$:
 \begin{equation}
    {{\mathbf{c}}_{\mathbf{j}}} = {[{c_{j1}},{c_{j2}}, \ldots ,{c_{jn}}]^{T}},
\end{equation}
where each element represents the corresponding  $k$-gram feature extracted by the filter ${{\mathbf{W}}_j}$.

A limitation for ${\mathbf{c_{j}}}$ is that it only contains features extracted by one filter ${{\mathbf{W}}_j}$.
It is necessary to introduce multiple filters to learning different features of $k$-gram in the text.
To overcome this limitation, we apply multiple filters (with different window sizes) to obtain complementary features.
All these feature vectors are assembled into a feature matrix:
 \[{\mathbf{C}} = [{{\mathbf{c}}_1},{{\mathbf{c}}_2}, \ldots ,{{\mathbf{c}}_i}, \ldots ,{{\mathbf{c}}_l}] \in {\mathbb{R}^{n\times l}},\]
 where ${{\mathbf{c}}_i}$ corresponds activations for the filter ${{\mathbf{W}}_i}$ on
 all the $k$-grams within the text; $l$ is the total number of filters.

Notably, the filters should not be treated equally since some of them are able to discover meaningful words or phrases while others may not. Thus, a trainable weight vector $\mathbf{w} \in \mathbb{R}^l $ is introduced for differing effects of various filters. The importance of each word can be obtained through a linear combination of rows within matrix ${\mathbf{C}}$. The result is normalized in order to keep stable at a reasonable scale. The result, which is denoted as ${\mathbf{a}}$, is computed as follows:
\begin{equation}
{\mathbf{a}} = softmax({\mathbf{C}}\mathbf{w}).
\end{equation}
Each dimension of ${\mathbf{a}}$ is viewed as the weight of a corresponding word.


Notably, the trained weight vector ${\mathbf{a}}$ usually attends to a specific
component of the text, like a special set of related words or phrases. It is
supposed to reflect one aspect, and/or component of the semantics of the text.
However, for practical text, there can be multiple components associated with its topic
or sentiment polarity. They together form
more than one aspect of the overall semantic meaning of the
text. So it is necessary to introduce multiple
weight vectors to learning the importance of each word from different aspects, as one vector focuses
on a special part and reflects one aspect of semantics contained in the text.
We give short movie reviews from the  Movie Review (MR) dataset
\cite{pang2005seeing} as examples:

\begin{figure}[!tbhp]
    \centering
    \fbox{
        \begin{minipage}{0.95\linewidth}
            \begin{itemize}
                \item saddled with \textbf{an unwieldly cast} of characters and angles , but the payoff is \textbf{powerful and revelatory}.

                \item It's \textbf{a brilliant, honest performance} by Nicholson, but the film is \textbf{an agonizing bore} except when the fantastic Kathy Bates turns up bravado Kathy!
            \end{itemize}
        \end{minipage}
    }
    \label{cases_multiple_comp}
    \caption{Examples of the Movie Review dataset.}
\end{figure}

These movie reviews are labeled with positive polarity and negative polarity, respectively.
In the texts, ``powerful and revelatory'' and ``a brilliant, honest performance'' express the positive sentiment,
while the ``an unwieldly cast'' and ``an agonizing bore'' do
negative sentiment. For text classification,
only focusing on the positive or negative sentiment part can mislead
sentiment polarity of the text.
A single weight vector ${\mathbf{a}}$ is usually
unable to concentrate on all the related components, especially for long sentences.

To tackle this problem, we adopt multiple hops of attention
to  represent the overall semantics of the text\cite{lin2017structured}:
$\mathbf{w}$ is replaced by a matrix ${{\mathbf{W}}} \in
{\mathbb{R}^{l \times z}}$ and the weight matrix ${\mathbf{A}}$ is  obtained by:
\begin{equation}
{\mathbf{A}} = \operatorname{softmax}({\mathbf{C}}{{\mathbf{W}}}),
\end{equation}
where $\mathbf{A} \in {\mathbb{R}^{n \times z}}$; dimension $z$ is hyper-parameter referring
to how many different weight vectors, $\operatorname{softmax}()$ is
applied along each column of the matrix. In other words, we use $z$ different linear
combinations of feature vectors to generate multiple weight vectors.

\subsection{Word's contextual representation learning}
The contexts help us to obtain a more precise word meaning.
In our model, we use bi-directional RNN to capture context information of each word and utilize a neural tensor layer
to compute the word's contextual representation.

RNN treats a text as a sequence and analyzes it as a data stream.
In our implementation, we adopt a variant of RNN named Gated Recurrent Units (GRU) \cite{cho2014learning}
to capture the contextual information around the focus word. The hidden state of the $t$-th word is
denoted as ${{\mathbf{h}}_t} \in \mathbb{R}^m$  and  computed as follows:

%
\begin{equation}
\begin{split}
{{\mathbf{z}}_t} &= \sigma ({{\mathbf{W}}_z}{{\mathbf{x}}_t} + {{\mathbf{U}}_z}{{\mathbf{h}}_{t - 1}}), \\
{{\mathbf{r}}_t} &= \sigma ({{\mathbf{W}}_r}{{\mathbf{x}}_t} + {{\mathbf{U}}_r}{{\mathbf{h}}_{t - 1}}),\\
{{\mathbf{\tilde h}}_t} &= \tanh ({{\mathbf{W}}_h}{{\mathbf{x}}_t} + {{\mathbf{r}}_t} \circ {{\mathbf{U}}_h}{{\mathbf{h}}_{t - 1}}),\\
{{\mathbf{h}}_t} &= (1 - {{\mathbf{z}}_t}) \circ {\mathbf{{\tilde{h}}}_t} + {{\mathbf{z}}_t} \circ {{\mathbf{h}}_{t - 1}},
\end{split}
\end{equation}
where ${{\mathbf{x}}_t}$, ${{\mathbf{h}}_{t - 1}}$ are the input;
${{\mathbf{h}}_t}$ is the output; parameters ${{\mathbf{W}}_z},{{\mathbf{W}}_r},{{\mathbf{W}}_h} \in {\mathbb{R}^{m \times d}}$, ${{\mathbf{U}}_z},{{\mathbf{U}}_r},{{\mathbf{U}}_h} \in {\mathbb{R}^{m \times m}}$. $\sigma(\cdot)$, $\tanh (\cdot)$ and $\circ$ refer to element-wise sigmoid, hyperbolic tangent functions and multiplication respectively.

Gated recurrent units use update gate ${{\mathbf{z}}_t}$ and reset gate
${{\mathbf{r}}_t}$ for the aim of controlling the flow of information. This controllability enables it to capture long-term dependencies. Update gate ${{\mathbf{z}}_t}$ influences the weights of
previous hidden state  ${{\mathbf{h}}_{t-1}}$ and new memory
$\mathbf{\tilde{h}}_{t}$ on generating ${{\mathbf{h}}_t}$. Reset gate
${{\mathbf{r}}_t}$ determines how much information from ${{\mathbf{h}}_{t-1}}$
is kept to generate $\mathbf{\tilde{h}}_{t}$. Update gate ${{\mathbf{z}}_t}$ and
reset gate ${{\mathbf{r}}_t}$ are computed using ${{\mathbf{h}}_{t - 1}}$ and
${{\mathbf{x}}_t}$.

In order to capture context information of each word,
we use bi-directional GRU \cite{bahdanau2014neural} to process the text ${X}$. The bi-directional GRU contains the forward $\overrightarrow {\operatorname{GRU}}$ which processes the text ${{\mathbf{x}}}_{1:n}$ and backward $\overleftarrow {\operatorname{GRU}}$ which processes the text from ${{\mathbf{x}}_{n:1}}$:
    \begin{align}
        \overrightarrow {{{\mathbf{h}}_t}}  = \overrightarrow {\operatorname{GRU}} ({{\mathbf{x}}_t},{{\mathbf{h}}_{t - 1}}),\\
        \overleftarrow {{{\mathbf{h}}_t}}  = \overleftarrow {\operatorname{GRU}} ({{\mathbf{x}}_t},{{\mathbf{h}}_{t + 1}}).
    \end{align}
We treat ${{\mathbf{h}}_0}$ and ${{\mathbf{h}_{n+1}}}$ as zero vectors in our method.

Hidden states $\overrightarrow {{{\mathbf{h}}_t}}$ and $\overleftarrow {{{\mathbf{h}}_t}}$
contains contextual information around $i$-th word in forward and backward directions respectively.
Conventionally, two vectors are concatenated and fed into a fully connected network
since none of them contains the full contextual information.
However, following this methodology, these two vectors do not have influence over each other which causes the
compositional meaning of context is missing. We argue the word's contextual representation is
more precise when the context information from different directions is collected and fused.
It would be plausible to add a powerful composition function such that two vectors could interact directly and explicitly.
To this end, a neural tensor layer \cite{socher2013recursive} is utilized to fuse the forward vector and the backward vector, which allows for directly, possibly more flexible interaction between them.
Unlike the fully connected network where input vectors only interact implicitly,
a neural tensor layer permits greater interactions among each element of hidden states.
It is a powerful tool to serve as a composition function for our purpose.
We hope it aggregates meanings from forward and backward hidden states more accurately than simply concatenating individuals does.
Concretely, we take  $\overrightarrow {{{\mathbf{h}}_t}}$ and $\overleftarrow{{{\mathbf{h}}_t}}$ as neural tensor layer's input each time step and compute $[\widehat{h_{t}}]_i $ as follow:
\begin{equation}
[{\widehat h_{t}}]_i = \tanh ({\overrightarrow{\mathbf{h}}_t}{\mathbf{V}^i}{\overleftarrow{\mathbf{h}_t}} + {b_i}),
\end{equation}
where ${\mathbf{V}^i} \in {\mathbb{R}^{m \times m}}$, $b_i \in {\mathbb{R}}$. Through neural tensor layer, each element in $[{\widehat h_{t}}]_i$ can be viewed
as a different type of the intersection between $\overrightarrow {{{\mathbf{h}}_t}}$ and $\overleftarrow {{{\mathbf{h}}_t}}$.
In our model, $\widehat{\mathbf{h}}$
has the same size as $\overrightarrow {{{\mathbf{h}}_t}}$ and $\overleftarrow {{{\mathbf{h}}_t}}$.
After getting $\widehat{\mathbf{h}}_t$,
we concatenate $\overrightarrow {{{\mathbf{h}}_t}}$, $\overleftarrow {{{\mathbf{h}}_t}}$ and $\widehat{\mathbf{h}}_t$ together to obtain vector
$\overleftrightarrow{{{\mathbf{h}}_t}}$:
\begin{equation}
{\overleftrightarrow{\mathbf{h}}_t} = [\overrightarrow{\mathbf{h}_t}^T, \overleftarrow{\mathbf{h}_t}^T, \widehat{\mathbf{h}}_t]^T,
\end{equation}
which summarizes the context information  centered around ${{\mathbf{x}}_t}$.

For the sake of convenience, we stack all$\overleftrightarrow{\mathbf{h}}_t$ into matrix ${\mathbf{H}}$ whose shape is $n$-by-$3m$:
\begin{equation}
{\mathbf{H}} = [\overleftrightarrow{\mathbf{h}_1};\overleftrightarrow{\mathbf{h}}_2; \ldots ;\overleftrightarrow{\mathbf{h}}_n]
\end{equation}

\subsection{Text representation learning}

After attention matrices ${\mathbf{A}}$ and ${\mathbf{H}}$ are obtained, we get the text representation ${\mathbf{S}}$ by multiplying transpose of ${\mathbf{A}}$ to ${\mathbf{H}}$:
\begin{equation}
{\mathbf{S}} = {{\mathbf{A}}^T}{\mathbf{H}}
\end{equation}
Each row of ${\mathbf{S}}$ is a linear combination of hidden states. Then we
concatenate all rows in ${\mathbf{S}}$ and pass them into a fully connected
$ \operatorname{softmax} $ layer to get prediction  over various classes.
We use cross-entropy loss as the objective function to train our model:
\begin{equation}
L(\mathbf{y}, \mathbf{{\hat{y}}}) = -\sum_{i}^{} {y_{i}}\log{\hat{y_{i}}},
\end{equation}
where $\hat{y_{i}}$ is prediction and $y_{i}$ is true probability of class $i$.

We choose dropout \cite{hinton2012improving} and L2 regularization in order to prevent overfitting.
Dropout prevents overfitting by randomly omitting a subset of features at each iteration of a training procedure.
For our case, instead of omitting each element independently,
dropout is applied on matrix ${\mathbf{C}}$ column-wisely for our model, i.e. instead of omitting each element independently, a random number of columns of ${\mathbf{C}}$ are set to zeros in the training procedure. That is, only a proportion of filters will be updated. We also carry out dropout on matrix ${\mathbf{H}}$ and ${\mathbf{S}}$. L2 regularization is added to all parameters with the same regularization force. Training is done through stochastic gradient descent using adam optimizer \cite{kingma2014adam} with learning rate $r=0.001$.

\section{Experimental Analysis}
\subsection{Datasets}
Six different datasets are used in the experiments for evaluating our method. The experiments involve various types of tasks, including sentiment analysis, news categorization and topical classification. All the datasets can be obtained from Internet freely and some have been treated as benchmark data in a range of research tasks. For clarity, the descriptions of these datasets are listed as follows:

\textbf{MR}: Movie reviews are labeled by their sentiment polarity \cite{pang2005seeing}, which is a binary classification \cite{chen-2009-pcvm,chen-2014-epcvm}.

\textbf{Yelp}:
Yelp challenge round 10 consists of about 4.6 million yelp review-star pairs.
We randomly select 120k from each class to form our dataset.

\textbf{Ag's news}:
The dataset used in our experiment is the same as in \cite{zhang2015character}, containing the top 4 largest classes.
The number of training and testing samples for each class is 30000 and 1900.

\textbf{20 Newsgroups}: It consists of about 20,000  documents across 20 different newsgroups.
We select four major categories (comp, politics, rec, and religion) as \cite{hingmire2013document} and \cite{lai2015recurrent}.

\textbf{Sogou news}: This dataset is obtained from SogouCA news corpora\footnote{http://www.sogou.com/labs/resource/ca.php}. 
It contains 
18 different Chinese topic channels. We choose 10 major classes as a multi-class classification \cite{Lyu-2019-mpcvm} and include 6500 samples randomly from each class.

\textbf{Yahoo! Answers}:
This dataset\cite{zhang2015character} is  used for topic classification.
It contains 10 classes and each class contains 140k training samples and 5k testing samples.

\subsection{Experimental Settings}

\begin{table*}[!htbp]
    \centering
    \caption{The descriptions of datasets. }
\begin{tabular}{lcccccc}
    \toprule
    Dataset        & Classes & Train set & Dev set & Test set & Avg. text length & Classification Task         \\ \midrule
    MR             &    2    &     10662     &          -          &      CV      &               20 & Sentiment analysis          \\
    Yelp           &    5    &    500000     &        50000        &    50000     &              134 & Sentiment analysis          \\
    AG’s news      &    4    &    108000     &        12000        &     7600     &               42 & English news categorization \\
    20Newsgroups   &    4    &     7521      &         835         &     5563     &              429 & English news categorization \\
    Sogou news     &   10    &     50000     &        5000         &    10000     &              350 & Chinese news categorization \\
    Yahoo! Answers &   10    &    1260000    &       140000        &     5000     &              112 & Topic classification        \\ \bottomrule
\end{tabular}
\label{datasets}
\end{table*}

We preprocess datasets as follows.
 For all corpus, we retain words with regards to their frequencies and replace those less frequent as \texttt{<UNK>}. The English corpora are format to lower case. For English corpus, we use the publicly available \emph{word2vec} vectors\footnote{https://code.google.com/p/word2vec/} trained on 100 billion words from Google news to initialize our word representations. Words are initialized to random vector if they are absent in the set of pre-trained words. For Chinese corpus, we use \emph{jieba}\footnote{https://github.com/fxsjy/jieba} to segment the sentences. We pre-train 50-dimensional word vectors on Chinese \emph{wikipedia} using \emph{word2vec}. All word vectors are updated during training.

For MR dataset, we perform 10-fold cross-validation to evaluate our method suggested by \cite{kim2014convolutional}. The other datasets are separated into training set, development set, and test set. The detailed statistics of datasets and separations are listed in Table \ref{datasets}. For comparison convenience, the evaluation metric of the 20Newsgroups is selected to be the Macro-F1 measure suggested by \cite{hingmire2013document} and \cite{lai2015recurrent}. For other datasets, classification accuracy is used as the metric.

The hyperparameters setting is as follows: We set the size of the hidden state of GRU for all the datasets as $m=50$ and the total number of filters are set as $l=300$. The dropout rate is set to 0.5.
We search the other hyperparameters in a wide range and choose those that yield the highest accuracy in the development set. The hyperparameters searching ranges are listed in Table \ref{hyperparameters}.

\begin{table}
    \centering
    \begin{threeparttable}
        \caption{Searching ranges of the hyperparameters. }
        \setlength{\tabcolsep}{8.8mm}{\begin{tabular}{lr}
            \toprule
            \multicolumn{1}{c}{Parameter} &   \multicolumn{1}{c}{Searching Ranges} \\ \midrule
            $ m $\tnote{1}                & \{\{1\}, \{3\}, \{1,2,3\}, \{3,4,5\}\} \\
            $z$\tnote{2}                  &                  \{5, 15, 20, 25, 30\} \\
            $ \lambda $\tnote{3}          &               \{0.0001, 0.002,0.005 \} \\
            $l$\tnote{4}                  &             \{30, 150, 300, 450, 600\} \\ \bottomrule
        \end{tabular}}
        \begin{tablenotes}
            \item[1]{Filter size, single filter, and multiple filters are both considered}
            \item[2]{Number of different weight vectors}
            \item[3]{Weight of $ l2 $ regularization}
            \item[4]{Number of filters}
        \end{tablenotes}
        \label{hyperparameters}
    \end{threeparttable}
\end{table}

\subsection{Comparison of methods}
We compare our method with a variety of text classification methods:

\textbf{SVM + Unigram/Bigrams}: Support Vector Machine (SVM) is traditionally used as a baseline in  sentiment analysis and/or text categorization. It  has shown that SVM using unigram and bigram as text features could be a strong performer, even in some sophisticated scenarios \cite{wang2012baselines}.

\textbf{CNN}: A simple CNN with one layer of convolution proposed in \cite{kim2014convolutional} also included in comparison algorithms. It performs remarkably well in the sentence-level classification.

\textbf{Bi-GRU}: As a prototype of RNN, Bi-directional GRU is chosen as a baseline algorithm.  In its implementation, for comparison purposes, we perform the mean-pooling along with its hidden states and pass the output through a fully connected softmax layer. This step is analogous in our method to obtain the text representation.

\textbf{Self-attentive LSTM}: Multiple-hop of attention to LSTM is introduced to get sentence embedding. Experimental results in \cite{lin2017structured} show that this method achieves significant performance gain compared against its baseline methods.

\textbf{RCNN}: Lai \textit{et al.} propose recurrent convolution neural network(RCNN) for text classification\cite{lai2015recurrent}. This model uses a recurrent structure to capture contextual information as far as possible. The authors claim this holistic structure may introduce considerable less noise than in some local features. Finally, a max-pooling layer is appended to construct the representation of the text.

\textbf{ClassifyLDA-EM}: LDA-based methods perform well in capturing the semantics of texts.  Among them, ClassifyLDA-EM has served as one typical paradigm \cite{hingmire2013document} so we select it as a comparison in the LDA-based methods.


\textbf{MV-RNN}: Matrix-vector recursive neural network(MV-RNN)\cite{socher2012semantic} assigns a vector and a matrix to every node in a parse tree of the sentence.
A representation of the sentence is computed bottom-up by recursively combining the words according to the syntactic structure of a parse tree.

\textbf{Char-conv}: Zhang \textit{et al.} first apply convolutional networks (Char-conv) only on the character level \cite{zhang2015character}.
The results show Char-conv is an effective method for text classification.

\textbf{VDCNN}: Conneau \textit{et al.} propose very deep convolutional neural networks(VDCNN) which process texts at the character level \cite{conneau2017very}. We choose this model as another comparison with  character-level CNN.

\textbf{TCNN-SM}: Wang and Deng propose a novel tightly-coupled convolutional neural network with spatial-temporal memory (TCNN-SM).
It comprises feature-representation and memory functional columns\cite{wang2017tightly}. We choose this model as another comparison with CNN.

\textbf{Region.emb}: 
This method \cite{qiao2018anew} first calculates the vector representation of the whole $k$-gram through the interaction between adjacent words in the text,
then adds all the vector representations of $k$-gram to get the whole text representation.

\textbf{ID-LSTM}: Zhang \textit{et al.} use reinforcement learning to simplify sentences in texts. They retain only words with rich semantics
and delete words that are not useful for classification tasks\cite{zhang2018learning}.
The experimental results show that this method performs well in different text classification tasks.

\begin{table*}[!thp]

        \centering
        \begin{tabular}{lcccccc}
            \toprule
                                                       &       MR       &      Yelp      &   Ag's news    &  20Newsgroups  &  Yahoo!answer  &   Sogou news   \\ \midrule
            SVM + Unigram $\dagger$                    &     76.92      &     56.63      &     91.78      &     93.05      &     71.32      &     81.70      \\
            SVM + Bigram  $\dagger$                    &     78.08      &     59.10      &     91.70      &     93.23      &     70.75      &     84.62      \\
            CNN          $\dagger$                     &     80.87      &     65.11      &     92.82      &     95.60      &     72.45      &     84.41      \\
            Bi-GRU        $\dagger$                    &     80.61      &     64.76      &     92.41      &     95.72      &     73.91      &     84.81      \\
            RCNN               $\dagger$               &     81.13      &     65.89      &     92.50      &     95.81      &     74.17      &     85.04      \\
            Self-attentive LSTM   $\dagger$            &     81.20      &     65.83      & \textbf{93.08} & \textbf{96.43} &     74.30      &     84.99      \\ \midrule
            MV-RNN $\ddagger$ \cite{socher2012semantic}&      79.0      &       -        &       -        &       -        &       -        &       -        \\
 ClassifyLDA-EM $\ddagger$ \cite{hingmire2013document} &       -        &       -        &       -        &     93.60      &       -        &       -        \\
        Char-conv $\ddagger$ \cite{zhang2015character} &       -        &       -        &     91.45      &       -        &     71.20      &       -        \\
            VDCNN $\ddagger$ \cite{conneau2017very}    &       -        &       -        &     91.33      &       -        &     73.43      &       -        \\
            TCNN-SM $\ddagger$ \cite{wang2017tightly}  &       -        &       -        &     92.01      &       -        &       -        &       -        \\
            Region.emb $\ddagger$ \cite{qiao2018anew}  &       -        &       -        &      92.8      &       -        &      73.7      &       -        \\
            ID-LSTM $\ddagger$ \cite{zhang2018learning}&      81.6      &       -        &      92.2      &       -        &       -        &       -        \\ \midrule
            CRNN w/o NTL                               &     81.80      &     66.18      &     92.54      &     96.13      &     74.66      &     85.20      \\
            CRNN                                       & \textbf{82.03} & \textbf{66.51} &     93.05      &     96.31      & \textbf{75.05} & \textbf{85.55} \\ \bottomrule
        \end{tabular}
        \caption{The classification accuracy on several benchmark data sets. The best result on each data is marked in bold.\label{tab:3}. Note that ``CRNN w/o NTL'' means
        CRNN is implemented without the neural tensor layer (NTL). $\dagger$  marks results re-implemented in this paper. $\ddagger$ marks results reported in the original paper. }
        \label{classification comparison}
\end{table*}

\section{Results and Discussion}
\subsection{Classification Comparison}
Results of our method and comparison methods are listed in Table \ref{classification comparison}.

1. Except  Sogou news dataset, the performance of neural network methods on word-level is better than traditional methods based on bag-of-words features (SVM + Unigram/Bigram, ClassifyLDA-EM). The better performance can be explained as follows: compared with traditional methods that treat texts as unordered sets of $n$-grams, the neural network methods can easily capture more contextual information and get a holistic view on the text. However, for Sogou news dataset, neural networks only give marginal advantage in comparison with SVM + Bigram. The possible reason is due to a lack of well-trained word vectors.

2. Our model outperforms CNN, TCNN-SM, and Bi-GRU on all datasets.
CNN can capture $k$-gram information from text. However, the length of the extracted segments is limited by the window size and causes its failure in capturing long term dependency.
TCNN-SM equips CNN with memory functional column to retain memories of different granularity and fulfills selective memory for historical information.
Bi-GRU could also capture long term dependency in the text.
This poor performance of TCNN-SM and Bi-GRU is partly caused by its equal treatment of words.
TCNN-SM uses the aggregation vector of the final time step to represent the text,
which does not take the difference of contribution of each word into consideration\cite{wang2017tightly}.
Rather than treating words differently as we do in our method, Bi-GRU performs mean-pooling to get representations of texts.
This step does not take into consideration the relative importance between each word, messing semantically important words into a collection of unimportant ones. In contrast, our method captures local information with CNN to obtain their relative importance. To demonstrate this point, we remove the neural tensor layer from our architecture but keep others unchanged. The comparisons is carried out on all datasets. The experimental results in Table \ref{classification comparison} demonstrate that using $k$-gram information as weights is critical for improvement.


3. ID-LSTM and Self-attentive LSTM use different ways to extract the words that contribute more semantically in the text.
Except Ag’s news and 20Newsgroups datasets, our model performs better than ID-LSTM and Self-attentive LSTM. This performance gain may be achieved by adding the neural tensor layer for capturing the surrounding contextual information around each word. This neural tensor layer brings compositional information between forward and backward RNN in a direct way. The experimental results in Table \ref{classification comparison} confirm what we expect. In the current experimental setting, the method with neural tensor layer beats its counterpart trained without it. 

4. RCNN also uses recurrent and convolution structure in its model. RCNN utilizes recurrent structure to learn preceding and post contexts, and then concatenates them with word embedding to represent each word. After feeding each word representation to a fully connected neural network, it uses a max-pooling layer to get a fixed-length vector
for the text. Although the max-pooling layer succeeding in finding the most important latent semantic factors in the document\cite{lai2015recurrent}, it does not explicitly learn the  importance of a word. Compared with their model, CRNN uses convolution operation to learn the importance of each word from different aspects and utilizes the neural tensor layer to fuse preceding and post context information. Table \ref{classification comparison} shows our model outperforms RCNN in these datasets.

5. Compared with Char-conv and VDCNN, our model has better performance on Ag's news and Yahoo!answer dataset. The probable explanation is that our model could benefit from pre-trained word vectors that capture meaningful semantic regularities.

6. We compare CNN, Bi-GRU and our model with MV-RNN on MR dataset and experimental results show that convolution-based and recurrent-based model has better performance.
One reason is that MV-RNN captures contextual information using semantic composition under a parse tree, which may introduce noise during tree construction.
Another reason is that MV-RNN does not leverage the advantages of pre-trained word vectors\cite{socher2012semantic}.

%
%
%

\subsection{Results with Varying Hyperparameters}
We exploit our method's performance with various filter region sizes in different classification tasks. In our experiment, we explore the effect of the filter size using one region size and different region sizes. For a fair comparison, the total number of filters are the same. We report the best accuracy on the development set of each dataset in Table \ref{filter_sizes}.
From Table \ref{filter_sizes} it can be concluded that, for the topic classification task, the performance of our method is insensitive to the choice of window sizes. On the contrary, for sentiment analysis, a large window size usually results in  high performance. This is partially due to the fact that sentimental sentences always have more adjectives/adverbs to enhance its expressiveness or for emphasis.

\begin{table}[!hthp]
    \centering
    \caption{Accuracies on the development sets of three datasets with different filter sizes.}
    \begin{tabular}{ccccc}
        \toprule
        Filter size &   Ag's news    &      Yelp      &   Yahoo!answer   &  \\ \midrule
           \{1\}    &     92.75      &     65.57      &     78.44      &  \\
           \{3\}    &     92.69      &     66.03      &     78.45      &  \\
        \{1, 2, 3\} & \textbf{92.96} &     66.14      & \textbf{78.50} &  \\
        \{3, 4, 5\} &     92.81      & \textbf{66.53} &     78.26      &  \\ \bottomrule
    \end{tabular}
    \label{filter_sizes}
\end{table}

\subsection{Case study}

\begin{figure*}[!htbp]
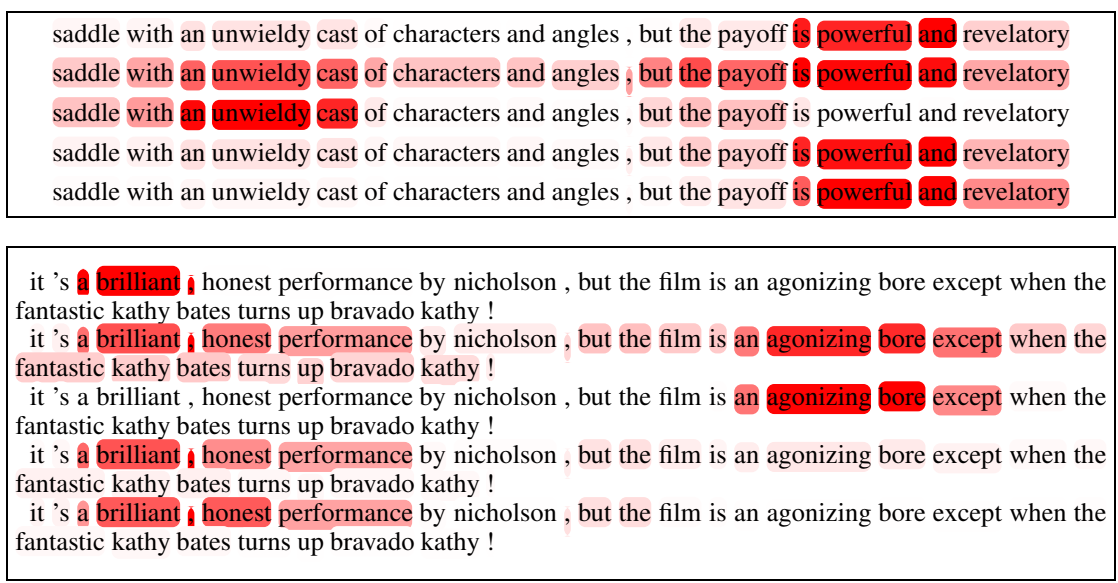

    \centering
    \subfigure{
        \fbox{%
            \parbox[c][2.5cm][c]{14.5cm}{
                \centering
                \subfigure{
                    \mybox[3.615015162375658]{saddle} \mybox[3.959876774243483]{with} \mybox[11.292622173669583]{an} \mybox[10.899617729332292]{unwieldy} \mybox[10.759658069182324]{cast} \mybox[1.8870472710356123]{of} \mybox[0.6781343229033441]{characters} \mybox[0.6353729742905245]{and} \mybox[0.2736358315655454]{angles} \mybox[0.5645716212116559]{,} \mybox[1.2555966069445545]{but} \mybox[8.244052224279121]{the} \mybox[8.618818613974168]{payoff} \mybox[89.46479733126647]{is} \mybox[86.02688851759565]{powerful} \mybox[100.0]{and} \mybox[18.27991893055195]{revelatory}
                }
                \hspace{.1in}
                \subfigure{
                    \mybox[21.21057914107418]{saddle} \mybox[34.86450525638781]{with} \mybox[61.68662729731227]{an} \mybox[67.45549470088072]{unwieldy} \mybox[60.898896741584494]{cast} \mybox[34.35937820583125]{of} \mybox[23.22600242175673]{characters} \mybox[23.191842084038008]{and} \mybox[18.904892365309845]{angles} \mybox[36.26175747583773]{,} \mybox[48.59580143114919]{but} \mybox[70.20762914126247]{the} \mybox[59.48162221655202]{payoff} \mybox[100.0]{is} \mybox[97.30818277847695]{powerful} \mybox[95.04153754214286]{and} \mybox[34.33692306310607]{revelatory}
                }
                \hspace{.1in}
                \subfigure{
                    \mybox[24.009327163067613]{saddle} \mybox[39.089972783968825]{with} \mybox[97.43406608696746]{an} \mybox[100.0]{unwieldy} \mybox[85.48275755049856]{cast} \mybox[8.446553036066492]{of} \mybox[1.67500441187659]{characters} \mybox[1.914774034634072]{and} \mybox[1.131301199453047]{angles} \mybox[1.943712434125766]{,} \mybox[16.100880388147253]{but} \mybox[25.519215105128623]{the} \mybox[24.603416372027258]{payoff} \mybox[9.954671993526828]{is} \mybox[0.5638194355890489]{powerful} \mybox[1.063531567057502]{and} \mybox[0.6836091731079239]{revelatory}
                }
                \hspace{.1in}
                \subfigure{
                    \mybox[4.5976753227651015]{saddle} \mybox[5.728534776168636]{with} \mybox[12.286996437301106]{an} \mybox[10.88630077279765]{unwieldy} \mybox[10.73644125275516]{cast} \mybox[3.6251653000144892]{of} \mybox[3.055738497669787]{characters} \mybox[3.1801618181499762]{and} \mybox[2.56262973247057]{angles} \mybox[6.6063926590401625]{,} \mybox[9.208434820215706]{but} \mybox[19.58577179627696]{the} \mybox[18.767550384619646]{payoff} \mybox[85.82641111064346]{is} \mybox[93.8285951768121]{powerful} \mybox[100.0]{and} \mybox[29.06284440077147]{revelatory}
                }
                \hspace{.1in}
                \subfigure{
                    \mybox[1.2561799638824713]{saddle} \mybox[1.411469709419101]{with} \mybox[3.8115518960924084]{an} \mybox[3.3937785900603625]{unwieldy} \mybox[3.4123633224738144]{cast} \mybox[0.8521795183861116]{of} \mybox[0.489852792870694]{characters} \mybox[0.42243640733602733]{and} \mybox[0.31308672045737196]{angles} \mybox[1.843376096670367]{,} \mybox[2.080609831809841]{but} \mybox[7.6460765715775985]{the} \mybox[10.674772022876803]{payoff} \mybox[63.76469207226617]{is} \mybox[100.0]{powerful} \mybox[99.07813752296174]{and} \mybox[45.61408896063414]{revelatory}
                }
            }%
        }
    }

    \subfigure{
        \fbox{%
            \centering
            \parbox[c][4.2cm][c]{14.5cm}{%

                \mybox[0.0]{ } \mybox[0.0036012199298182442]{it} \mybox[0.004325006664408247]{'s} \mybox[99.6030083916837]{a} \mybox[99.61396872356654]{brilliant} \mybox[100.0]{,} \mybox[0.40358094657475463]{honest} \mybox[0.3905087450239887]{performance} \mybox[0.006308511386052974]{by} \mybox[0.003726047117873598]{nicholson} \mybox[0.01619961452539936]{,} \mybox[0.03816218567815174]{but} \mybox[0.040916009691736366]{the} \mybox[0.0446238601114748]{film} \mybox[0.02438056301402157]{is} \mybox[0.07493789098865265]{an} \mybox[0.09749254319093473]{agonizing} \mybox[0.09809417797184725]{bore} \mybox[0.047603477643679136]{except} \mybox[0.008701078084127721]{when} \mybox[0.015906026238047995]{the} \mybox[0.023377322185022344]{fantastic} \mybox[0.02670970300407525]{kathy} \mybox[0.018642977908451236]{bates} \mybox[0.008564100022545158]{turns} \mybox[0.008198831632881374]{up} \mybox[0.010624961675622286]{bravado} \mybox[0.010934781161398156]{kathy} \mybox[0.0072676420664803155]{!}

                \mybox[0.0]{ } \mybox[3.721522444369816]{it} \mybox[5.393530906997785]{'s} \mybox[56.54010372522651]{a} \mybox[68.9838704070162]{brilliant} \mybox[100.0]{,} \mybox[51.18633370994638]{honest} \mybox[38.209909208925964]{performance} \mybox[8.207640410768812]{by} \mybox[7.367830280818165]{nicholson} \mybox[16.085638978464768]{,} \mybox[23.504337600589803]{but} \mybox[25.048830206483192]{the} \mybox[20.629078773445787]{film} \mybox[20.966277522823724]{is} \mybox[52.60710913524602]{an} \mybox[84.83950916623095]{agonizing} \mybox[83.57585815690993]{bore} \mybox[55.41706390274847]{except} \mybox[20.705407964057272]{when} \mybox[16.202036792616383]{the} \mybox[19.15993966139373]{fantastic} \mybox[19.257859442210403]{kathy} \mybox[21.08993578211808]{bates} \mybox[14.84305654231688]{turns} \mybox[17.404819674402194]{up} \mybox[16.36402128845663]{bravado} \mybox[13.789185474583181]{kathy} \mybox[7.1861199896965005]{!}

                \mybox[0.0]{ } \mybox[0.002945712266409347]{it} \mybox[0.002945712266409347]{'s} \mybox[0.0031890663353641544]{a} \mybox[0.002928873721603939]{brilliant} \mybox[0.003033379915135666]{,} \mybox[0.0022033484812608263]{honest} \mybox[0.002753856926346373]{performance} \mybox[0.0028139855152045706]{by} \mybox[0.002641347945597866]{nicholson} \mybox[0.003150880464860044]{,} \mybox[0.045459400712590685]{but} \mybox[0.04604024222712783]{the} \mybox[0.0606530699678528]{film} \mybox[0.835945049371041]{is} \mybox[54.19017585057543]{an} \mybox[98.92149297745976]{agonizing} \mybox[100.0]{bore} \mybox[46.74781934249444]{except} \mybox[2.002019517373403]{when} \mybox[0.10643215455681326]{the} \mybox[0.005594612856811737]{fantastic} \mybox[0.0070669461994926795]{kathy} \mybox[0.033507567725596044]{bates} \mybox[0.0370487796884075]{turns} \mybox[0.08755157617177875]{up} \mybox[0.06128140697848071]{bravado} \mybox[0.05594675150268349]{kathy} \mybox[0.002139382968601042]{!}

                \mybox[0.0]{ } \mybox[2.1416738047981645]{it} \mybox[3.0732221464380927]{'s} \mybox[59.48435112466055]{a} \mybox[67.51069615218316]{brilliant} \mybox[100.0]{,} \mybox[44.508087917309844]{honest} \mybox[35.394273989460864]{performance} \mybox[2.2482548811805634]{by} \mybox[1.2891152810182558]{nicholson} \mybox[7.676037200815824]{,} \mybox[11.537193708255725]{but} \mybox[12.323620135503841]{the} \mybox[6.850765212145182]{film} \mybox[4.354463537300948]{is} \mybox[7.0674123469709516]{an} \mybox[8.94791271803143]{agonizing} \mybox[8.184688749659658]{bore} \mybox[5.2367163683055695]{except} \mybox[2.605609086443749]{when} \mybox[3.011499004276542]{the} \mybox[5.60856669211592]{fantastic} \mybox[5.822763159299755]{kathy} \mybox[5.875618456388402]{bates} \mybox[3.144599001073141]{turns} \mybox[3.3568810364066426]{up} \mybox[3.3053093683429395]{bravado} \mybox[2.801310107127035]{kathy} \mybox[1.6351165203335818]{!}

                \mybox[0.0]{ } \mybox[4.640084663288192]{it} \mybox[5.18361223330693]{'s} \mybox[41.27472467203339]{a} \mybox[65.0386706076328]{brilliant} \mybox[100.0]{,} \mybox[64.5439238472443]{honest} \mybox[36.511539521248956]{performance} \mybox[1.0766768380447755]{by} \mybox[0.44184052591515843]{nicholson} \mybox[12.235679985955136]{,} \mybox[12.702349928620766]{but} \mybox[13.243058641535498]{the} \mybox[1.6328417794788508]{film} \mybox[1.2768063323792687]{is} \mybox[0.8091120654342986]{an} \mybox[0.6069540443599714]{agonizing} \mybox[0.45895699113702015]{bore} \mybox[0.4837447881488245]{except} \mybox[0.53007379368518]{when} \mybox[0.9622034595814167]{the} \mybox[2.936184962651095]{fantastic} \mybox[3.0838777364970884]{kathy} \mybox[2.793850492271771]{bates} \mybox[0.7146054821197753]{turns} \mybox[0.4954747032329636]{up} \mybox[0.8461230365789459]{bravado} \mybox[0.830638107567961]{kathy} \mybox[0.6337503341290591]{!}
            }%
        }
    }
    \caption{Two test samples from MR dataset. The first one is labeled as ``positive" and the second one is labeled as ``negative"}
    \label{case_vis}
\end{figure*}
In our method, the relative importance of words is computed in the matrix ${\mathbf{A}}$. Each column of ${\mathbf{A}}$ reweighs the contributions of each word representation to the final text representation.

With multiple weights vectors, our method may concentrate on different words or components that are most relevant to the current classification task.
In order to validate that our model is able to select informative words in a text,
We take two test samples from MR dataset and visualize the matrix ${\mathbf{A}}$ for illustration in Fig \ref{case_vis}.
Note that these samples are labeled as ``positive" and ``negative", respectively and both of them have words with opposite sentiment polarity.
To make the weights more clear, we set a unique color map for each sentence, in which the thickness of color denotes the magnitudes of weight.
From Fig \ref{case_vis}, it can be observed that whatever the polarity of examples are
 both positive and negative key words can be extracted by our method,
 reflecting that it may capture multiple semantically meaningful components, including opposite ones.


\begin{table}[!hbtp]
    \centering
    \caption{Sampled important words of different classes from Ag's news dataset.}
    \begin{tabular}{cccc}
        \toprule
           World    &   Sports   &  Business  &    Sci/Tech     \\ \midrule
          missile   &   bryant   &  worldcom  &      dell       \\
         venezuela  &   knicks   & resources  &      hdtv       \\
          olympic   & volleyball & federation & microprocessors \\
        afghanistan &    team    &  prepaid   &      sony       \\
         infantry   &  batsman   & greenback  &    infrared     \\
           greek    &   boxing   &  earnings  &      ipod       \\
          african   &   soccer   & capitalism &     symbian     \\
           japan    &    mvp     &    coal    &    adapters     \\
         european   &   hockey   &   stock    & supercomputing  \\
         oklahoma   &  baseman   & insurance  &    bluetooth    \\ \bottomrule
    \end{tabular}
    \label{AG key words}
\end{table}

Our method also works well for news categorization. We take Ag's news dataset as an example. In this experiment, we randomly select 10 different important words in the test set of Ag's news in Table \ref{AG key words}.
The important words are the ones that have the maximum of each weight vector in the text.
The results demonstrate that the matching of words and classes is good in our method.


\section{Conclusion}
In this paper, we propose convolutional recurrent neural networks
for text classification. In our model, we use a convolutional neural
network to compute the weight matrix which shows the relative
importance of each word from different aspects. Meanwhile, we use
the bi-directional RNN to process text and introduce neural a tensor
layer to fuse forward and backward context information to obtain the
word's contextual representation. Finally, the weight matrix and the
word's contextual representation are combined to get the text
representation. The experimental results indicate that the
proposed method not only achieves better performance compared with
previous methods, but also provides insight into which words in the
text contribute to the classification decision which can be of value
in applications and analysis. In the future, we want to explore 
features \cite{jiang-2019-probabilistic} extracted from the method to learning  
methods, such as model space \cite{chen-2013-model,chen-2014-learning}.

%

\bibliographystyle{IEEEtran}
\bibliography{bibtex}

\end{document}